\documentstyle[12pt,epsbox]{article}
\input{standard92.mac}
\input{local.mac}
\input{symbol.mac}

\def\OMIT#1{}

\begin{document}

\begin{tabbing}
{\bf Technical Reports on Mathematical and Computing Sciences:} TR-C123\\
{\bf title:}~
Practical Algorithms for On-line Sampling\\
{\bf authors:}~
Carlos Domingo$^1$, Ricard Gavalda$^1$, and Osamu Watanabe$^2$\\
{\bf affiliation:}\\
1.~\=\kill
1.\>Dept.\ de LSI,
    Univ.\ Politecnica de Catalunya\\
  \>Campus Nord, Modul C5, 08034-Barcelona, Spain.\\
  \>{\bf email:} $\{$carlos, gavalda$\}$@lsi.upc.es\\
2.\>Dept.\ of Mathematical and Computing Sciences,
    Tokyo Institute of Technology\\
  \>Meguro-ku Ookayama, Tokyo 152-8552.\\
  \>{\bf email:} watanabe@is.titech.ac.jp\\
{\bf acknowledgements to financial supports:}\\
1.\>Supported in part by ESPRIT LTR Project no.\ 20244 - ALCOM-IT\\
  \>and CICYT Project TIC97-1475-CE.\\
2.\>Supported in part by the Ministry of Education, Scinece, Sports
    and Culture,\\
  \>Grant-in-Aid for Scientific Research on Priority Areas
    (Discovery Science), 1998.
\end{tabbing}

\noindent
{\bf Abstract.}~
One of the core applications of machine learning to knowledge discovery
consists on building a function (a hypothesis) from a given amount of
data (for instance a decision tree or a neural network) such that we
can use it afterwards to predict new instances of the data. In this
paper, we focus on a particular situation where we assume that the
hypothesis we want to use for prediction is very simple,
and thus,
the hypotheses class is of feasible size. We study the problem
of how to determine which of the hypotheses in the class is almost the
best one. We present two on-line sampling algorithms for selecting
hypotheses, give theoretical bounds on the number of necessary
examples, and analize them exprimentally. We compare them with the
simple batch sampling approach commonly used and show that in most of
the situations our algorithms use much fewer number of examples.

\section{Introduction and Motivation}

The ubiquity of computers in business and commerce has lead to
generation of huge quantities of stored data. A simple commercial
transaction, phone call or use of a credit card is usually stored in a
computer. Todays databases are growing in size and therefore there is
a clear need for automatic tools for analyzing and understanding these
data. The field known as knowledge discovery and data mining aims at
understandings and developing all the issues concern with the
extraction of patterns from vast amount of data. Some of the
techniques used are basically machine learning techniques. However,
due to the restriction that the data available is very large, many
machine learning techniques do not always scale well and can not just
simply be applied.

One of the core applications of machine learning to knowledge
discovery consists of building a function from a given amount data
(for instance a decision tree or a neural network) such that we can
later use it to predict the behavior of new instances of the
data. This is commonly know as concept learning or supervised
learning.

Most of the previous research in machine learning has focused on
developing efficient techniques for obtaining {\em highly} accurate
predictors.  For achieving high accuracy, it is better that learning
algorithms can handle complicated predictors, and developing efficient
algorithms for complicated predictors has been studied intensively in
machine learning.

On the other hand, for knowledge discovery, there are some other
aspects of concept learning that should be considered, and we discuss,
in this paper, one of them.  We study concept learning (or, more
simply, hypotheses selection) for a particular situation that we
describe in the following.  We assume that in our situation we have a
class $\HC$ of very simple hypotheses, and we want to select one of
the reasonably accurate hypotheses from them, by using a given set of
data, i.e., labeled examples.  Since hypotheses we deal with are very
simple, we cannot hope, in general, to find highly accurate hypotheses
in $\HC$. On the other hand, the size of hypotheses space $\HC$ is
relatively small and feasible.  We also assume that the size of the
data available is huge, and thus, it is very inefficient to use all
examples in the dataset.  Simple hypotheses have been studied before
by several researchers and it has been reported that in some cases
they can achieve surprisingly high accuracy (see,
e.g.,~\cite{WGT90,Hol93,AHM95}).  Moreover, with the new discover of
voting methods like boosting~\cite{FS95}, bagging~\cite{Bre96b}, or
error-correcting output codes~\cite{DB95}, several of these hypotheses
can be combined in a way that the overall precision becomes extremely
high.

Perhaps the paper by Holte~\cite{Hol93} best exemplifies our
problem. In that paper he performs several experiments with some 
datasets from the repository of the University of California at
Irvine. His learning algorithm is extremely simple, just obtains a
training set from the datasets, it builds a set of very simple
hypotheses according to the different features of the dataset (see the
paper for more details on how to build the set of simple hypotheses)
and then selects the hypothesis that has the highest accuracy on the
training set. It turns out that this simple approach is indeed
efficient since for most of the datasets the accuracy is between 80
and 90 percent.  His choice of training set size is totally arbitrary,
$2/3$ of the whole dataset. If the dataset avalaible is huge as it
happens in many situations then this choice might be very inefficient.

On the other hand, the obvious approach for solving this problem that
is commonly used in computational learning theory~\cite{Val84} is to
first choose randomly a certain number $m$ of examples from the
dataset, and then select the hypothesis that performs best on these
examples.  (We will call this simple hypotheses selection {\em Batch
Selection} (BS) in this paper.)  The number $m$ is calculated so that
the best hypotheses on the selected sample is close to the real best
one with high probability; such $m$ can be calculated by using uniform
convergence bounds like the Chernoff or the Hoeffding bound (see,
e.g.,
\cite{KV94} for some examples of this approach).
However, if we want to apply this method in a real setting we will
encounter two problems. First, the theoretical bounds are usually too
pessimistic and thus the bounds obtained are not practical. Second, to
obtain this bounds we need to have certain knowledge about the
accuracy of hypotheses in a given hypothesis space.  What is usually
assumed is that we know a lower bound on the accuracy of the best
hypothesis. Again, this lower bound might be far from the real
accuracy of the best hypothesis and thus the theoretical bound becomes
too pessimistic.  Or even worst, in many applications we just do not
know anything about the accuracy of the hypotheses.

In this paper we propose two algorithms for solving this problem,
obtain theoretical bounds of their performance, and evaluate them
experimentally.  Our goal is to obtain algorithms that are useful in
practice but that also have certain theoretical guarantees about their
performance.  The first distinct characteristic is that we obtain the
examples in an on-line manner rather than in batch. The second is that
the number of examples has less dependency on the lower bound of the
accuracy than the above obvious Batch Selection.  More specifically,
if $\gammaz$ is the accuracy of the best hypothesis, and $\gamma$ is
the lower bound for $\gammaz$ we would use, then the sample size $m$
for Batch Selection given by the theoretical bound is ${\cal
O}(1/\gamma^2)$ (ignoring dependencies in other parameters).  On the
other hand, the sample size of our first algorithm is ${\cal
O}(1/\gamma\gammaz)$, and that of the second one is ${\cal
O}(1/\gammaz^2)$.

The paper is organized as follows. In the following section we give
some definitions. In Section~3 we state the two selection
algorithms and prove their performance theoretically.
In the last section we compare and analyze them experimentally.

\section{Preliminaries} 

Throughout this paper, we use $\HC$ and $n$ to denote the set of
hypotheses and its size, and use ${\cal D}$ to denote a distribution
on instances.
We assume some $\EX$ that generates instances
according to the distribution ${\cal D}$,
and each selection algorithm can make use of $\EX$.
For any $h\in\HC$, let $\prdh$ denote the accuracy of
$h$, that is, the probability that $h$ gives a collect prediction to
$x$ for a randomly given $x$ under the distribution ${\cal D}$.  Let
$\hz$ denote the best hypothesis in $\HC$ (w.r.t.${\cal D}$); that is,
$\prdhz$ $=$ $\max\{\prdh | h\in \HC\}$.  Let $\gammaz$ denote
$\prdhz-1/2$; that is, $\prdhz$ $=$ $1/2+\gammaz$.

We use $\rhou$ and $\rhol$ to denote upper and lower tail
probabilities of independent Bernoulli trials.  More specifically, for
any $t\ge1$ and $p$, $0\le p\le1$, consider $t$ independent random
variables $X_1,...,X_t$ each of which takes 0 and 1 with probability
$1-p$ and $p$.  Then for any $\eps>0$, we define $\rhou(p,\eps,t)$ and
$\rhol(p,\eps,t)$ as follows:

\[
\rhou(p,\eps,t)
~=~\Pr\{\,\sum_{i=1}^t X_i>pt+\eps t\,\},{\rm~~and~~}
\rhol(p,\eps,t)
~=~\Pr\{\,\sum_{i=1}^t X_i<pt-\eps t\,\}.
\]

For these tail probabilities, several bounds have been used in the
literature; here we make use of the following ones
(see, e.g., \cite{KV94}).

\begin{theo}\label{theo:hb}
(Hoeffding bound)\\
For some constant $\chernoff>0$,
and for any $p$, $\eps$, and $t$,
we have
\[
\rhou(p,\eps,t)~<~\exp(-\chernoff\eps^2t),{\rm~~and~~}
\rhol(p,\eps,t)~<~\exp(-\chernoff\eps^2t).
\]

\remark
The Hoeffding bound used in the literature uses $\chernoff=2$.
Later in this paper,
we will use different constants
that work respectively in a certain situation.
\end{theo}

By using this bound, we can estimate the sufficient number of examples
to guarantee that Batch Selection, the simple hypothesis selection
algorithm, yields a hypothesis of reasonable accuracy with high
probability.  (In the following, we use $\BS(\delta,\gamma, m)$ to
denote the execution of Batch Selection for parameters $\delta$,
$\gamma$ and $m$, the sample size.  Recall that the hypotheses space, its
size, and the accuracy of best hypothesis is fixed, throughout this
paper, to $\HC$, $n$, and $1/2+\gammaz$.)

\begin{theo}\label{theo:BS}
For any $\gamma$ and $\delta$, $0<\gamma,\delta<1$, if
$\gamma\le\gammaz$ and $m=16\ln(2n/\delta)/(\chernoff\gamma^2)$ then
with probability more than $1-\delta$, $\BS(\gamma,\delta,m)$ yields
some hypothesis $h$ with $\prdh$ $\ge$ $1/2+\gammaz/2$.
\end{theo}

\beginproof
Follows from the Hoeffding bound in Theorem~\ref{theo:hb}. 
\endproof

\section{On-line Selection Algorithms and Their Analysis}

Here we present our two on-line selection algorithms
and investigate their reliability and efficiency theoretically.
In our analysis of the algorithms
we count each while-iteration as one step;
thus, the number of steps is equal to
the number of examples needed in the algorithm.
By ``at the $t$ step''
we precisely mean ``at the point just after the $t$th while-iteration.''
Throughout this section,
we denote by $\#_t(h)$ the number of examples for which the
hypothesis $h$ succeeds within $t$ steps.
It will be also useful for our analysis to partition the
hypothesis space in two sets depending on the precision of each
hypothesis. Thus, let $\HCge$ (resp., $\HCless$) denote the set of
hypotheses $h$ such that $\prdh$ $\ge$ $1/2+\gammaz/2$ (resp., $\prdh$
$<$ $1/2+\gammaz/2$). This partition can be done in an arbitrary way. The
complexity of our algorithms depends on it but can be easily adapted
to a more restrictive condition (for instance $h\in\HCge$ if $\prdh\ge
1/2+3\gammaz/4$) if it is needed for a particular
application. Obviously, the more demanding is the definition of $\HCge$,
the greater is the complexity of our algorithms.

In our analysis,
we ignore small difference occurring
by taking ceiling or floor function,
or by computing real number with finite precision.

\subsection{Constrained Selection Algorithm}
 
We begin by introducing a function that is used to determine an
important parameter of our algorithm.  For a given $n$, $\delta$, and
$\gamma$, define $\bCS(n,\delta,\gamma)$ by
 
\[
\bCS(n,\delta,\gamma)
~=~
{16\over{\chernoff\gamma^2}}\cdot
\ln\left(\,
   \left({2n\over\delta}\right)
   \left({16e\over\chernoff(e-1)\gamma^2}\right)\,\right)
~=~
{16\over{\chernoff\gamma^2}}\cdot
\ln\left({32en\over\chernoff(e-1)\delta\gamma^2}\right).
\]
 
\noindent
Now our first algorithm, that we denote by $\CS$ from constrained
selection, is stated as follows.
 
\begin{algo}
$\CS(\delta,\gamma)$\\
\>$B$ $\plet$ $3\gamma \bCS(n,\delta,\gamma)/4$;\\
\>set $w(h)$ $\plet$ $0$ for all $h\in\HC$;\\
\>{\bf while}~$\forall h\in\HC$ $[\,w(h)<B\,]$~{\bf do}\\
\>\>$(x,b)$ $\leftarrow$ $\EX$;\\
\>\>$\HC'$ $\plet$ $\{\,h\in\HC$\,:\,$h(x)=b\,\}$;~~$n'$ $\plet$ $|\HC'|$;\\
\>\>{\bf for each}~$h\in\HC$~{\bf do}\\
\>\>\>{\bf if}~$h\in\HC'$~\={\bf then}~
      \=$w(h)$ $\plet$ $w(h)+1-n'/n$;\\
\>\>\>                     \>{\bf else}
      \>$w(h)$ $\plet$ $w(h)-n'/n$;\\
\>\>{\bf end-for}\\
\>{\bf end-while}\\
\>output $h\in\HC$ with the largest $w(h)$;
\end{algo}
 
Note that the number $n'$ of successful hypotheses may vary at each
step, which makes our analysis difficult.  For avoiding this
difficulty, we approximate $n'$ as $n/2$; that is, we assume that a
half of hypotheses in $\HC$ always succeeds on a given example.  In
other words, we assume the following.

\beginsome{Assumption.}
After $t$ steps (i.e., after $t$ while-iterations),
the following holds for each $h\in\HC$.

\[
w(h)~=~\#_t(h)-t/2,
\]

\remark
In fact, we can modify $\CS$ to the one satisfying this assumption;
that is, use a fixed, i.e., 1/2, decrement term instead of $n'/n$.  As
our experiments show, both algorithms seem to have almost the same
reliability, while the modified algorithm has more stable complexity.
We believe, however, that the original algorithm is more efficient in
many practical applications.  (See the next section for our experiments
and discussion.)
\endsome

First we investigate the reliability of this algorithm.

\begin{theo}\label{theo:CS}
For any $\gamma$ and $\delta$, $0<\gamma,\delta<1$, if
$\gamma\le\gammaz$, then with probability more than $1-\delta$,
$\CS(\gamma,\delta)$ yields some hypothesis $h\in\HCge$.
\end{theo}

\beginproof
We estimate the error probability $\Perr$, i.e., the probability
that $\CS$ chooses some hypothesis with $\prdh$ $<$ $1/2+\gammaz/2$,
and show that it is less than $\delta$, in the following way.

\[
\begin{array}{lcl}
\Perr
&=&
\displaystyle
\Pr_{\CS}\{~
\bigcup_{t\ge1}
[\,\mbox{$\CS$ stops at the $t$th step
         and yields some $h\in\HCless$}\,]~\}\\[2mm]
&\le&
\displaystyle
\Pr_{\CS}\{~
\bigcup_{t\ge1}
[\,
\exists h\in\HCless
[\,\mbox{$w(h)$ reaches  $B$
         at the $t$th step (for the first time)}\,]\\[-3mm]
& &~~~~~~~~~~~\land~
\forall h\in\HCge
[\,\mbox{$w(h)$ has not reached $B$ within $t-1$ steps}\,]\,]~\}
\\[2mm]
\OMIT{&\le&
\displaystyle
\sum_{h\in\HCless}
\Pr_{\CS}\{~
\bigcup_{t\ge1}
[\,[\,\mbox{$w(h)$ reaches $B$
            at the $t$th step (for the first time)}\,]\\[-3mm]
& &~~~~~~~~~~~~~~~~~~~~\land~
[\,\mbox{$w(\hz)$ has not reached $B$ within $t-1$ steps}\,]\,]~\}
\\[2mm]
}&\le&
\displaystyle
\sum_{h\in\HCless}
\Pr_{\CS}\{~
\bigcup_{t\ge1}~
[\,[\,\mbox{$w(h)$ reaches $B$ within $t$ steps}\,]\\[-3mm]
& &~~~~~~~~~~~~~~~~~~~~~~\land~
[\,\mbox{$w(\hz)$ has not reached $B$ within $t-1$ steps}\,]\,]~\}.
\end{array}
\]

Let $\tzt=\bCS(n,\delta,\gamma)$ and $\tz=(\gamma/\gammaz)\tzt$.
(Note that $\tz\le\tzt$.)  We estimate the above probability
considering two cases: $t\le\tz$ and $t\ge\tz+1$.  That is, we
consider the following two probabilities.

\[
\begin{array}{lcl}
P_1(h)
&=&
\displaystyle
\Pr_{\CS}\{~
\bigcup_{t\le\tz}~
[\,[\,\mbox{$w(h)$ reaches  $B$ within $t$ steps}\,]\\[-3mm]
& &~~~~~~~~~~~~~~\land~
[\,\mbox{$w(\hz)$ has not reached $B$ within $t-1$ steps}\,]\,]~\},
{\rm~~and}\\[2mm]
P_2(h)
&=&
\displaystyle
\Pr_{\CS}\{~
\bigcup_{\tz+1\le t}~
[\,[\,\mbox{$w(h)$ reaches $B$ within $t$ steps}\,]\\[-3mm]
& &~~~~~~~~~~~~~~~~\land~
[\,\mbox{$w(\hz)$ has not reached $B$ within $t-1$ steps}\,]\,]~\}.
\end{array}
\]

\noindent
In Lemma~\ref{lemm:lower} and Lemma~\ref{lemm:upper} below,
we prove that both $P_1(h)$ and $P_2(h)$ are bounded by
$\delta/2n$ for any $h\in\HCless$.
Therefore we have

\[
\Perr
~\le~
\sum_{h\in\HCless}P_1(h)+P_2(h)
~\le~
n\left({\delta\over 2n}+{\delta\over 2n}\right)
~=~\delta.
\]
\square

\begin{lemm}\label{lemm:lower}
For any $h\in\HCless$,
we have
$P_1(h)$ $\le$ $\delta/2n$.
\end{lemm}

\beginproof
We bound the probability
$P'_1(h)$ $=$
$\Pr_{\CS}\{\,\bigcup_{t\le\tz}\,[\,w(h)$
reaches to $B$ within $t$ steps\,$]\,\}$.
Clearly $P_1(h)\le P'_1(h)$.

The probability $P'_1(h)$ is in fact
the same as
the probability that
$w(h)$ reaches to $B$ in $\tz$ steps.
Now suppose that
$w(h)$ reaches to $B$ in $\tz$ steps.
Then for some $t\le\tz$,
$w(h)\ge B$ at the $t$th step
(i.e.,
just after the $t$th step).
From our assumption,
we have $w(h)$ $=$ $\#_t(h)-t/2$ at the $t$th step.
Also recall that
$B=3\gamma\tzt/4$
and that
$\Exp[\#_t(h)]$ $<$ $t/2+\gammaz t/2$
(since $h\in\HCless$).
Hence,

\[
\begin{array}{l}
\mbox{$w(h)\ge B$ at the $t$th step}\\
~\iff~
\#_t(h)-t/2~\ge~B~=~3\gamma\tzt/4~=~3\gammaz\tz/4\\
~\iff~
\#_t(h)~\ge~\Exp[\#_t(h)]+(t/2+3\gammaz\tz/4-\Exp[\#_t(h)])\\
~\imply~
\#_t(h)~>~\Exp[\#_t(h)]+(3\gammaz\tz/4-\gammaz t/2)
       ~>~\Exp[\#_t(h)]+\gammaz\tz/4.
\end{array}
\]

\noindent
Therefore,
if $w(h)$ reaches to $B$ within $\tz$ steps,
then $\#_t(h)>\Exp[\#_t(h)]+\gammaz\tz/4$ for some $t\le\tz$.
Hence,
by using the Hoeffding bound~\ref{theo:hb},
we can derive the following bound.
(Here recall that
$\gamma\le\gammaz$ and $\tz=(\gamma/\gammaz)\tzt$.)

\[
P'_1(h)
~\le~
\exp\left(-\chernoff\left({\gammaz\tz\over{4t}}\right)^2t\right)
~\le~
\exp\left(-{\chernoff\gammaz^2\tz\over16}\right)
~\le~
\exp\left(-{\chernoff\gamma^2\tzt\over16}\right).
\]

\noindent
On the other hand,
by our choice of $\tzt$ (i.e., $\bCS$),
we have
$\exp(-\chernoff\gamma^2\tzt/16)$ $<$ $\delta/2n$.
\endproof

\begin{lemm}\label{lemm:upper}
$P_2(h)~\le~\delta/2n$.
\end{lemm}

\beginproof
First we note the following.

\[
\begin{array}{lcl}
P_2(h)
&=&
\displaystyle
\Pr_{\CS}\{~
\bigcup_{\tz+1\le t}~
[\,[\,\mbox{$w(h)$ reaches to $B$ within $t$ steps}\,]\\[-3mm]
& &~~~~~~~~~~~~~~~~~\land~
[\,\mbox{$w(\hz)$ has not reached to $B$ within $t-1$ steps}\,]\,]~\}\\[2mm]
&\le&
\displaystyle
\Pr_{\CS}\{~
\bigcup_{\tz+1\le t}~
[\,\mbox{$w(\hz)$ has not reached to $B$ within $t-1$ steps}\,]~\}\\[2mm]
&\le&
\displaystyle
\sum_{\tz+1\le t}~
\Pr_{\CS}\{\,\mbox{$w(\hz)$ has not reached to $B$ within $t-1$ steps}\,\}.
\end{array}
\]

\noindent
Thus,
we estimate the probability
$P'_2(h,t)$ $=$
$\Pr_{\CS}\{$ $w(\hz)$ has not reached to $B$ in $t$ steps~$\}$,
for each $t\ge\tz$.

Here we modify $\CS$ slightly
(which we call $\CS'$)
so that it does not terminate
even if some of the weights reaches to $B$,
and let $w_t(h)$ denote
the weight of $h$ at the $t$th step in the execution of $\CS'$.
Note that
if $w(\hz)$ has not reached to $B$ in $\CS$ within $t$ steps
(including the $t$th step),
then $w_t(\hz)<B$ in $\CS'$.
On the other hand,
we have

\[
\begin{array}{lcl}
w_t(\hz)~<~B
&\iff&
\#_t(\hz)-t/2~<~B~=~3\gamma\tzt/4~=~3\gammaz\tz/4\\
&\iff&
\#_t(\hz)~<~\Exp[\#_t(\hz)]+(t/2+3\gammaz\tz/4-\Exp[\#_t(\hz)])\\
&\iff&
\#_t(\hz)~<~\Exp[\#_t(\hz)]+(3\gammaz\tz/4-\gammaz t)
         ~\le~\Exp[\#_t(h)]-\gammaz t/4.
\end{array}
\]

\noindent
Therefore,
if $w(\hz)$ has not reached to $B$ in $t$ steps in $\CS$,
then $\#_t(\hz)$ $<$ $\Exp[\#_t(h)]-\gammaz t/4$ in $\CS'$.
Hence,
by using the Hoeffding bound again,
we get $P'_2(h,t)$ $<$ $\exp(-\chernoff\gammaz^2t/16)$.

Now we estimate
$\sum_{\tz+1\le t}P'_2(h,t)$.
First for any $\Delta\ge0$,
consider $P'_2(h,\tz+\Delta)$.
From the above,
we have $P'_2(h,t)$ $<$ $\pz\cdot\exp(-(\chernoff\gammaz^2/16)\Delta)$,
where $\pz$ $=$ $\exp(-(\chernoff\gammaz^2\tz/16))$.
Hence,
if $\Delta\ge 16/\chernoff\gammaz^2$,
then $P_1(h,\tz+\Delta)<\pz\cdot e^{-1}$.
In general,
if $\Delta\ge k(16/\chernoff\gammaz^2)$,
then $P_1(\tz+\Delta)<\pz\cdot e^{-k}$.
Therefore we have\footnote{%
Precisely speaking,
the factor $8/\gammaz^2$ should be $\lceil 8/\gammaz^2\rceil$;
but the effect of the ceiling function is negligible,
we omit it for simplifying our discussion.}

\[
\begin{array}{lcl}
\displaystyle
\sum_{t\ge\tz}P'_2(h,t)
&=&
\displaystyle
\sum_{\Delta\ge0}P'_2(h,\tz+\Delta)\\
&\le&
\displaystyle
\pz\cdot{16\over\chernoff\gammaz^2}\cdot{1\over{1-e^{-1}}}
~<~
{\delta\over 2n}
\cdot{{\chernoff(e-1)\gamma^2}\over{16e}}
\cdot{{16e}\over{\chernoff(e-1)\gammaz^2}}
~\le~
{\delta\over 2n}.
\end{array}
\]

\noindent
(Note that
$\pz$ $\le$ $\exp(-(\chernoff\gamma^2\tzt/16))$,
which is less than $(\delta/2n)(\chernoff(e-1)\gamma^2/16e)$
by our choice of $\tzt$ (i.e., $\bCS$).)
\endproof

Though valid, our estimation of error probability is not tight, and it
may not give us a useful bound $B$ for practical applications.  Here
under a certain assumption (i.e., the independence of hypotheses), we
can derive a much better formula for computing $B$.

\begin{theo}\label{theo:CSbetter}
Consider a modification of $\CS$,
where we use the following definition for $\bCS$.

\[
\bCS(n,\delta,\gamma)~=~
{16\ln(2n/\delta)\over{\chernoff\gamma^2}}.
\]

\noindent
Assume that for any $h$ and $h'$, the correctness of $h$ on a randomly
given example $x$ is independent from that of $h'$.  (See the proof
below for the precise condition.)  Then we can show the same
reliability for $\CS$ as Theorem~\ref{theo:CS} for the modified
algorithm.
\end{theo}

\beginproof
It is easy to see that the new $\bCS$ is good enough for showing
Lemma~\ref{lemm:lower} (i.e., $P_1(h)\le\delta/2n$); on the other
hand, the proof of Lemma~\ref{lemm:upper} requires the previous
$\bCS$.  Thus, we do over the estimation of $P_2(h)$ again.

This time we bound $P_2$ as follows.

\[
\begin{array}{lcl}
P_2(h)
&=&
\displaystyle
\Pr_{\CS}\{~
\bigcup_{\tz+1\le t}~
[\,[\,\mbox{$w(h)$ reaches  $B$ within $t$ steps}\,]\\[-3mm]
& &~~~~~~~~~~~~~~~~~\land~
[\,\mbox{$w(\hz)$ has not reached  $B$ within $t-1$ steps}\,]\,]~\}\\[2mm]
&\le&
\displaystyle
\sum_{\tz+1\le t}~
\Pr_{\CS}\{~
[\,\mbox{$w(h)$ reaches $B$ within $t$ steps}\,]\\[-3mm]
& &~~~~~~~~~~~~~~~~~\land~
[\,\mbox{$w(\hz)$ has not reached $B$ within $t-1$ steps}\,]~\}
\end{array}
\]

Now we use our assumption, the independence of hypotheses;
more specifically,
we assume, for any $h\in\HCless$, that
$\Pr\{\,[\,w(h)$
reaches $B$ within $t$ steps\,$]$ $\land$
$[\,w(\hz)$
has not reached  $B$ within $t-1$ steps\,$]\}$
$=$
$\Pr\{[\,w(h)$
reaches $B$ within $t$ steps\,$]\}$
$\times$
$\Pr\{[\,w(\hz)$
has not reached $B$ within $t-1$ steps\,$]\}$.
Then from the above, we obtain the following bound.

\[
\begin{array}{lcl}
P_2(h)
&\le&
\displaystyle
\sum_{\tz+1\le t}~
\Pr_{\CS}\{\,\mbox{$w(\hz)$ has not reached $B$ within $t-1$ steps}\,\}
\\[-2mm]
& &~~~~~~~~~\times~
\Pr_{\CS}\{\,\mbox{$w(h)$ reaches  $B$ within $t$ steps}\,\}.
\end{array}
\]

On the other hand, we can show that, for any $t\ge\tz+1$,
$\Pr_{\CS}\{$\,$w(\hz)$ has not reached  $B$ within $t-1$
steps\,$\}$ $\le$ $\delta/2n$.
(See the proof of Lemma~\ref{lemm:upper}.)
Therefore, we have

\[
P_2(h)
~\le~
\sum_{\tz+1\le t}\,
{\delta\over{2n}}\times
\Pr_{\CS}\{\,\mbox{$w(h)$ reaches $B$ within $t$ steps}\,\}
~\le~
{\delta\over{2n}}.
\]
\square
\paragraphskip

It may be unlikely that $\hz$ is independent from {\em all} hypotheses
in $\HCless$.  We may reasonably assume, however, that for any
$h\in\HCless$, there exists some $h'\in\HCge$ such that $h$ and $h'$
are (approximately) independent, and our poof above works similarly
for such an assumption.  Thus, in most cases, we may safely use the
simplified version of $\bCS$, and we will use it in the following
discussion.

Next let us discuss the complexity of our algorithm $\CS$.  Here by
``complexity'', we mean the number of steps that $\CS(\delta,\gamma)$
needs to yield a hypothesis, or in other words, the number of examples
used to select a hypothesis.

Consider the execution of $\CS$ on some $\delta>0$ and
$\gamma\le\gammaz$.  It is easy to see that, after $t$ steps, the
weight of $\hz$ becomes $\gammaz t$ on average.  Thus, on average, the
weight reaches  $B$ in $B/\gammaz$ steps\footnote{%
Precisely speaking,
our argument is not mathematically correct, because we
estimate here $\min\{t|\Exp[w_t(\hz)]\ge B\}$, whereas what we need to
estimate is $\Exp[\min\{t|w_t(\hz)\ge B\}]$.}.  From this
observation, we may use the following function for the average
complexity of $\CS(\delta,\gamma)$.

\[
\tCS(n,\delta,\gamma,\gammaz)
~=~{B\over\gammaz}
~=~{12\ln(2n/\delta)\over{\chernoff\gamma\gammaz}}.
\]

\subsection{Adaptive Selection Algorithm}

In this section we give a different algorithm that does not use any
knowledge on the accuracy of the best hypothesis in the class
( recall that algorithm $\CS$ used the knowledge of a lower bound on
$\gamma_0$.). To achieve this goal, we modify the condition of the
while loop so it is changing adaptively according to the number of
examples we are collecting. We call the algorithm $AS$ from adaptive
selection. The algorithm is stated as follows.

\begin{algo}
$\AS(\delta)$ \\
\>$S \plet \emptyset$;~~t $\plet$ 0;~~$\epsilon \leftarrow 1/5$;\\
\>{\bf while}~$\forall h\in\HC$
              $[\,\#_t(h)\le t/2+5t\eps/2\,]$~{\bf do}\\
\>\>$(x,b)\leftarrow \EX$;\\
\>\>$S$ $\plet$ $S\cup\{(x,b)\}$;~~$t$ $\plet$ $t+1$;\\
\>\>$\eps$ $\plet$ $\sqrt{4\ln(3n/\delta)/(\chernoff t)}$;\\
\>{\bf end-while}\\
\>output $h\in\HC$ with the largest $\#_t(h)$;\\
\end{algo}

\remark
The condition of the while-loop is trivially satisfied until the
algorithm collects enough number of examples for $S$, i.e., $\|S\|$
$>$ $4\ln(3n/\delta)/(\chernoff (1/5)^2)$.  Thus, in practice, we
start the while-loop after obtaining $4\ln(3n/\delta)/(\chernoff
(1/5)^2)$ examples for $S$.
\endsome

Again we begin by investigating the reliability of this algorithm.

\begin{theo}\label{theo:AS}
For any $\delta$, $0<\delta<1$, with probability more than $1-\delta$,
$\AS(\delta)$ yields some hypothesis $h\in\HCge$.
\end{theo}

\beginproof
Our goal is to show that when the algorithm stops it outputs a
hypothesis $h\in\HCge$ with probability more than $1-\delta$.  That
is, we want to show the following probability is larger than
$1-\delta$.  

\[
\begin{array}{lcl}
\Pcorrect
&=&
\displaystyle
\Pr_{\AS}\{~
\bigcup_{t\ge1}
[\,
\mbox{$\AS$ stops at the $t$th step and yields some $h\in\HCge$}
\,]~\}\\[2mm]
&=&
\displaystyle
\sum_{t\ge1}
\Pr_{\AS}\{~
\mbox{$\AS$ stops at the $t$th step and yields some $h\in\HCge$}~\}\\[2mm]
&=&
\displaystyle
\sum_{t\ge1}
\Pr_{\AS}\{~
\mbox{$\AS$ yields some $h\in\HCge$}
\,|\,\mbox{$\AS$ stops at the $t$th step}~\}\\[-2mm]
\displaystyle
& &~~~~~\times~
\Pr_{\AS}\{~\mbox{$\AS$ stops at the $t$th step}~\}.
\end{array}
\]

\noindent Consider any $t\ge1$, and assume in the following that the algorithm
stops at the $t$th step, i.e., just after the $t$th while-iteration.
(Thus, we discuss here probability under the condition that $\AS$
stops at the $t$th step.)  Let $\eps_t$ and $S_t$ be the value of
$\eps$ and $S$ at the $t$th step.  Also let $h$ be the hypothesis that
$\AS$ yields; that is, $\#_t(h)$ is the largest at the $t$th step.

By our choice of $\eps_t$, we know that $t$ $=$
$4\ln(3n/\delta)/(\chernoff\eps_t^2)$, and thus,
by Lemma~\ref{lemm:AS} given below,
the following inequalities
hold with probability $>$ $1-\delta$.  

\[ \prdhz~\le~\prdh+\eps_t, {\rm~~and~~} |\prdh - \#_t(h)/t|~\leq~\eps_t/2. \]

From the second inequality, we have that $\#_t(h)\leq t(\eps_t/2+
\prdh)$, and since we know that $1/2+\gammaz$ $=$ $\prdhz$ $\ge$
$\prdh$, we get that $\#_t(h)\leq t/2+t\gammaz+t\eps_t/2$.  Moreover,
since the algorithm stopped, the condition of the while-loop is not
satisfied and thus, the following holds.

\[
t/2 +5t\eps_t/2 ~\le~\#_t(h)~\le~t/2+t\gammaz+t\eps_t/2
\]

\noindent
This implies that $\eps_t\le\gammaz/2$.  With this fact together with
the first inequality above (i.e., $\prdhz~\le~\prdh+\eps_t$ ), we can
conclude that $1/2 + \gammaz/2\leq\prdh $.

Therefore, for any $t\ge1$, we have $\Pr_{\AS}\{\,\AS$ yields some
$h\in\HCge\,|\,\AS$ stops at the $t$th step\,$\}$ $>$
$1-\delta$. This, together with the fact that
$\sum_{t\ge1}\Pr_{\AS}\{~\mbox{$\AS$ stops at the $t$th step}\}=1$
proves the theorem.
\endproof

\begin{lemm}\label{lemm:AS}
For a given $\eps$, $0<\eps\le1$, let $t$ $=$
$4\ln(3n/\delta)/(\chernoff\eps^2)$, and consider the point in the
execution of the algorithm just after the $t$th step.  Then for any
$h\in\HC$ such that $\#_t(h)\geq \#_t(h_0)$ we have

\[
\Pr_{\AS}\{\,
[\,\prdhz\le\prdh+\eps\,]
~\land~
[\,|\prdh-\#_t(h)/t|\le\eps/2\,]\,\}
~>~1-\delta.
\]
\end{lemm}

\beginproof
Fix any $h\in\HC$ such that $\#_t(h)\geq \#_t(h_0)$ and let $A(h)$ and
$B(h)$ denote the following conditions.

\[
\begin{array}{lcl}
A(h)
&\iff&
[\,\prdhz\le\prdh+\eps\,]~\land~[\,|\prdh-\#_t(h)/t|\le\eps/2\,],
{\rm~~and}\\
B(h)
&\iff&
[\, \prdhz  - \#_t(h_0)/t\le\eps/2  \,]~\land~[\,|\prdh-\#_t(h)/t|\le\eps/2  \,].
\end{array}
\]

We first show that $B(h)$ implies $A(h)$.
Notice that
$B(h)$ implies that

\[
[\,(\prdhz  - \#_t(h_0)/t) +( \#_t(h)/t - \prdh )\le\eps\,]
~\land~
[\,|\prdh-\#_t(h)/t|\le\eps/2  \,].
\]

\noindent
Rewriting we obtain that
 
\[
[\,(\#_t(h)/t-\#_t(\hz)/t)+(\prdhz-\prdh)\le\eps\,]
~\land~
[\, |\prdh-\#_t(h)/t|\le\eps/2 \,],
\]

\noindent
and since $\#_t(h)\geq \#_t(h_0)$,
it must hold that

\[
[\,\prdhz\le\prdh+\eps  \,]
~\land~
[\,|\prdh-\#_t(h)/t|\le\eps/2  ,],
\]

\noindent
which is condition $A(h)$.

Now we show that $\Pr_{\AS}\{\lnot B(h)\}$ $<$ $\delta$.  Thus, by the
union bound and the Hoeffding bound (Theorem~\ref{theo:hb}), the
probability over the choice of sample $S$ of size $t$ (which is the
same as the probability over the execution of $\AS$ until the $t$th
step) that there exists one $h\in\HC$ such that $B(h)$ does not hold
is less than $3n\exp(\chernoff(\eps/2)^2t)$, which is, by choice of
$t$, equal to $\delta$.  Then since $\lnot A(h)$ $\imply$ $\lnot
B(h)$, the lemma follows.
\endproof

Next we discuss the complexity of the algorithm.
Here we can prove the following bound.

\begin{theo}\label{theo:AStime}
For any $\delta$, $0<\delta<1$, with probability more than $1-\delta$,
$\AS(\delta)$ terminates within $64\ln(3n/\delta)/\chernoff\gammaz^2$
steps.
\end{theo}

\beginproof
Here we use the same notation as above.  Notice first that while we
are in the while-loop, the value of $\eps$ is always strictly
decreasing.  Suppose that at some step $t$, $\eps_t$ has became small
enough so that $4\eps_t<\gammaz$.  Then from Lemma~\ref{lemm:AS} (the
condition of the lemma always holds due to our choice of $\eps$), with
probability $>$ $1-\delta$, we have that 
$t(\prdh-\eps_t/2)$ $\leq$ $ \#_t(h)$, and $\prdhz -\eps_t$ $\le$ $\prdh$.
Putting these two inequalities together, we obtain that
$t/2+t\gammaz-t\eps_t - t\eps_t/2$ $\le$ $\#_t(h)$ (since $\prdhz$ $=$
$1/2+\gammaz$).  Since we assumed that $4\eps_t<\gammaz$, we can
conclude that $t/2+5t\eps_t/2<\#_t(h)$, and thus, the condition of the
loop is falsified.  That is, the algorithm terminates (at least) after
the $t$th while-iteration.

Recall that $\eps_t$ is defined to be
$\sqrt{4\ln(3|\HC|/\delta)/(\chernoff t)}$ at any step.  Thus, when we
reach to the $t$th step with $t$ $=$
$64\ln(3|\HC|/\delta)/(\chernoff\gammaz^2)$, then it mush hold that
$\eps_t$ $<$ $\gamma/4$, and by the above argument, the algorithm
terminates with probability larger than $1-\delta$.

\remark
Thus,
we use the following function
for our theoretical bound for the number of examples used by $\AS$.

\[
\tAS(n,\delta,\gamma)~=~
{16\ln(3|\HC|/\delta)\over{\chernoff\gamma^2}}.
\]
\endproof

Again this theoretical bound is not tight.
As we will see in the next section,
our experiments show that
the value of $\eps$,
when the algorithm stops,
is close to $\gammaz/2$ instead of $\gammaz/4$.
Thus,
the number of examples is much smaller than this theoretical bound.


\section{Experimental Evaluation of the Algorithms}

We first summarize three selection algorithms considered, and state
functions that bound the sufficient number of examples to guarantee,
in theory,
that the algorithm selects with probability $>$ $1-\delta$
a hypothesis $h$ with $\prdh$ $\ge$ $1/2+\gammaz/2$.
(Recall that we assume that a given hypothesis set
$\HC$ has some $h$ with $\prdh$ $\ge$ $1/2+\gammaz/2$.)

\begin{tabbing}
\hskip5mm\=\+\hskip8mm\=
Bound: $\tCS(n,\delta,\gamma)$
$=$ $12\ln(2n/\delta)/(\chernoff\gamma\gammaz)$ on worst case.~~\=\kill
$\bullet$
Batch Selection: $\BS(n,\delta,\gamma)$ (see Introduction)\\
\>Bound:
$\tBS(n,\delta,\gamma)$ $=$ $16\ln(2n/\delta)/(\chernoff\gamma^2)$ on worst case.
\>Condition: $\gamma\le\gammaz$.\\
$\bullet$
Constrained Selection: $\CS(n,\delta,\gamma)$\\
\>Bound:
$\tCS(n,\delta,\gamma)$
$=$ $12\ln(2n/\delta)/(\chernoff\gamma\gammaz)$ on average.
\>Condition: $\gamma\le\gammaz$.\\
$\bullet$
Adaptive Selection: $\AS(n,\delta)$\\
\>Bound:
$\tAS(n,\delta)$ $=$ $64\ln(3n/\delta)/(\chernoff\gammaz^2)$ on worst case.
\>Condition: None.
\end{tabbing}

Thus, for example, if we know $\gammaz$ and use it as $\gamma$, then
$\tBS(n,\delta,\gamma)$ examples are enough to guarantee $1-\delta$
confidence for $\BS$.
We compare these theoretical bounds with the numbers that
we obtained through experiments.

First we describe the setup used in our experiments.  We decided
to use synthetic data instead of real datasets so that we can
investigate our algorithms in a wider range of parameter values.  (In
future work we are planning to evaluate also them with real data.)

The common fixed parameters involved in our experiments are $\delta$,
the confidence parameter, and $n$, the number of hypotheses in $\HC$.
Notice that these two parameters are inside a logarithm in the above
bounds; thus, results are not really affected by modifying them.
\OMIT{\footnote{However, the running time of the algorithms is
polynomial in $n$ so in our experiments with bigger $n$ (we tried
up to 1000) the running time became much slower while the number of
examples remained almost unchanged.}.}  

In fact, we verified this experimentally, and based on those
results we set them to $18$ for $n$, and $0.01$ for $\delta$;
that is, we require confidence of $99\%$.  The other parameter is the
accuracy of the best hypothesis, which is specified by $\gammaz$.  In
our experiments the value of $\gammaz$ ranges from $0.04$ to $0.3$
with a increment of $0.01$ (that is, the accuracy of the best
hypothesis ranges from $54\%$ to $80\%$ with a increment of $0.4\%$)
and we have a total of 65 different values.  For each $\gammaz$, we
distributed the $18$ hypotheses in $9$ groups of $2$ hypotheses,
where the accuracy of hypotheses in each group is set
$1/2-\gamma$, $1/2-3\gamma/4$, ..., $1/2+3\gamma/4$, $1/2+\gamma$.
The choice of the distribution of hypotheses accuracy does not affect
the performance of neither $\BS$ nor $\AS$ (because their performance depends
only on the accuracy of the best hypothesis).  On the other hand, it
seems to affect the performance of $\CS$.  For this reason, we also
tried other distributions of the hypotheses accuracy for $\CS$.
For a random number generator,
we used one explained in \cite{Sed88}.

For each set of parameters, we generated a {\em success pattern}
for each hypothesis $h$.  A {\em success pattern} is a $0/1$ string of
1000 bits that are used to determine whether the hypothesis $h$
predicts correctly for a given example.  That is, to simulate the
behavior of $h$ on examples from $\EX$, we just draw a random number
$i$ between $1$ and $1000$, and decide $h$ predicts correctly/wrongly
on the current example if the $i$th bit of the success pattern is 1/0.
Finally, for every fixed setting of all the parameters, we run
this experiments 30 times, i.e., run each algorithm 30 times, and
averaged the results.  This is what is reflected on the graphs we have
throughout this section.

\paragraphskip\noindent
\underline{1.~The Tightness of Theoretical Bounds}

\noindent
Let us assume that we know the value of $\gammaz$, not just a lower
bound.  Then, from the bounds summarized first, one may think that,
e.g., $\CS$ is more efficient than $\BS$.  It turned out, however, it
is not the case.  Our experiment shows that the number of required
examples is similar among three algorithms, and the difference is the
tightness of our theoretical bounds. Of course, this is for the case
when $\gammaz$ is known, see the subsection below for a discussion on
this issue.

\begin{figure}
\begin{minipage}{0.47\textwidth}
\begin{center}
\psbox[width=\textwidth]{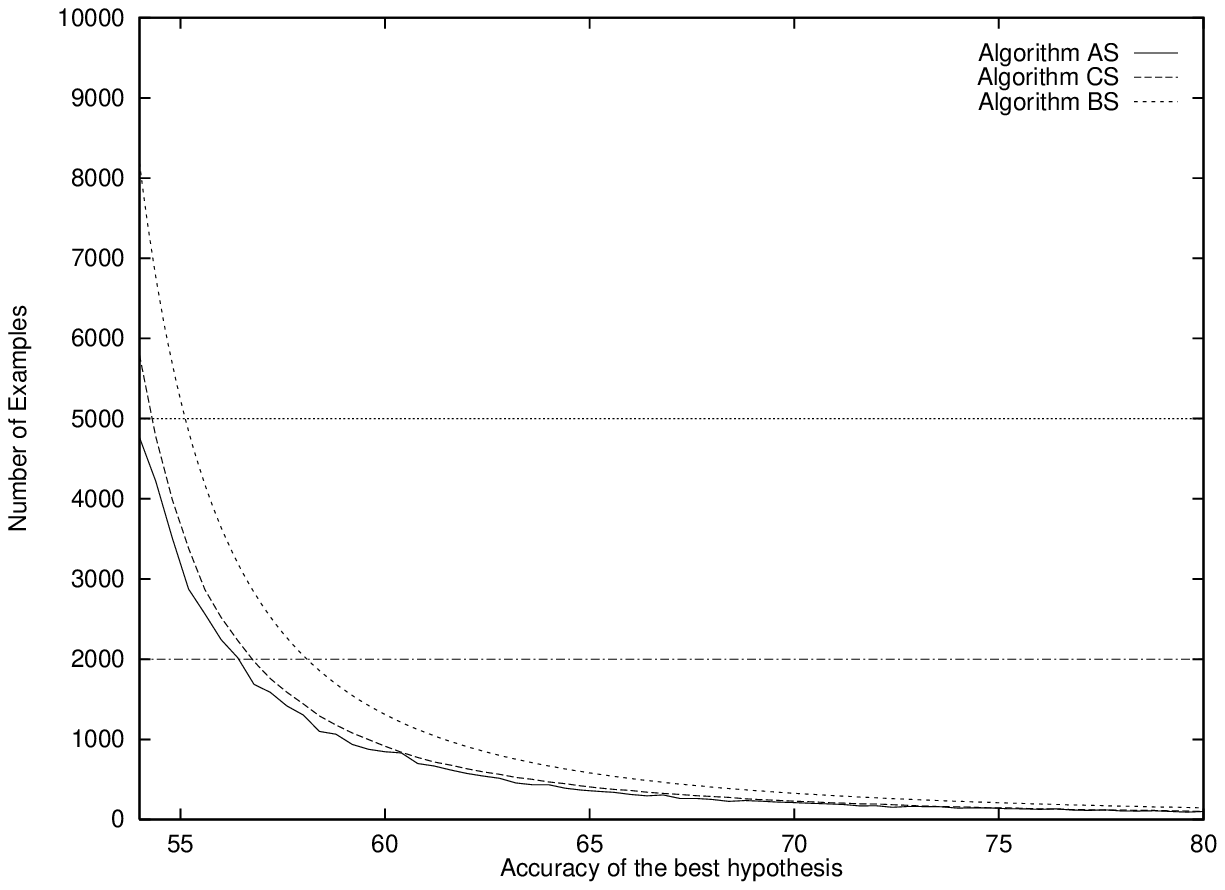}~\\
(a) With almost optimal constants.
\end{center}
\end{minipage}~
\begin{minipage}{0.47\textwidth}
\begin{center}
\psbox[width=\textwidth]{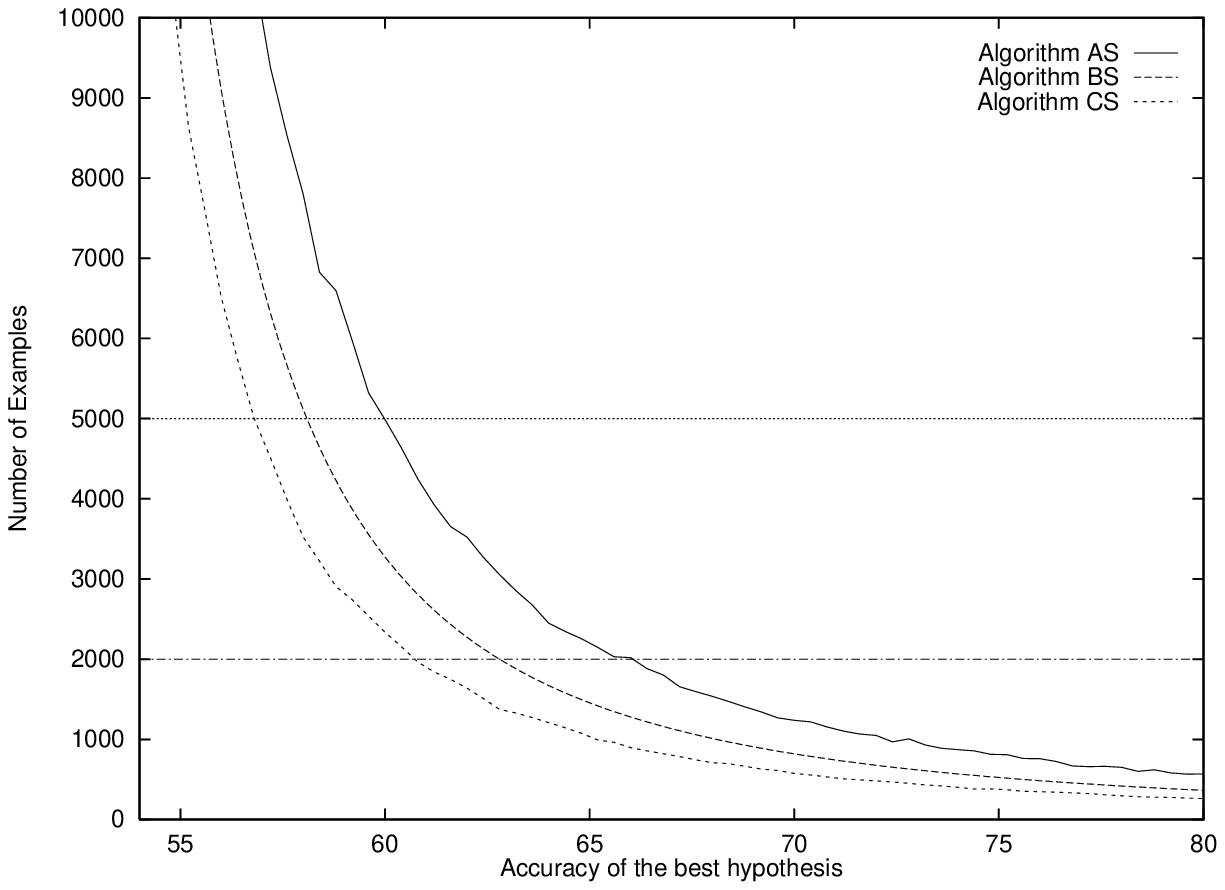}~\\
(b) With $\chernoff=4$.
\end{center}
\end{minipage}
\caption{the number of examples vs.\ $\gammaz$}
\label{fig:numofex}
\end{figure}

We checked that the ``necessary and sufficient'' number of examples is
proportional to $1/\gammaz^2$ (where $n$ and $\delta$ are fixed).
Thus, we changed the parameter $\chernoff$ to get the tightest bounds;
that is, for each algorithm, we obtained the smallest $\chernoff$ with
which the algorithm does not make any mistake in 30 runs.  The graph
(a) of Figure~\ref{fig:numofex} shows the number of examples needed by
three algorithms with such almost optimal constants.  There is not so
much difference, in particular, between $\CS$ and $\AS$.  Thus, the
tightness of our estimation seems to be the main factor of the difference of
theoretical bounds when $\gammaz$ is known. 

It is, however, impossible in real applications to estimate the
optimal constant and get the tightest bound.  Nevertheless, we can
still get a better bound by a simple calculation.  Recall that the
Hoeffding bound is a general bound for tail probabilities of Bernoulli
trials.  While it may be hard to improve the constant $\chernoff$ in
general, we can numerically calculate a better one for a given set of
parameters.  For instance, for our experiments, we can safely use
$\chernoff=4$ instead of $\chernoff=2$, and the difference is half;
e.g., $\tBS(18,0.01,0.1)$ (so the best hypothesis has $60\%$ of
accuracy) is 6550 with $\chernoff=2$ but 3275 with $\chernoff=4$.  The
graph (b) of Figure~\ref{fig:numofex} shows the number of examples
needed by three algorithms with $\chernoff=4$.  Thus, when using these
algorithms,  it is recommended to estimate first an appropriate
constant $\chernoff$, and use it in the algorithms.  For such usage,
$\CS$ is the most efficient for the set of parameters we used.

\paragraphskip\noindent
\underline{2.~Comparison of Three Algorithms}

\noindent
The graph (b) of Figure~\ref{fig:numofex} indicates that $\CS$ is best
(at least within this range of parameters) if $\gammaz$ or a good
approximation of it is {\em known}.  The situation differs a lot if we
do not know $\gammaz$.  For example, if $\gammaz=0.2$ but it is
underestimated as $0.05$, then $\BS$ and $\CS$ need 13101 and 2308
examples, while $\AS$ needs only 1237 examples; thus, in that case
$\AS$ is the most efficient.  This phenomenon is shown in Figure~2,
where we fixed $\gammaz$ to be $0.2\%$ (so the accuracy of the best
hypothesis is $70\%$), and we changed the value of the lower bound
$\gamma$ from $0.04$ to $0.2$.  Algorithm $\AS$ is not affected by the
value of $\gamma$, and hence it uses the same number of examples (the
horizontal line in the graph).  With this graph we can see that, for
instance, when $\gamma$ ranges from $0.04$ to $0.058$, algorithm $\AS$
is the most efficient, while from $0.058$ to $0.2$ algorithm $\CS$
becomes the best; but in any case, the difference is not so big within
this range of $\gamma$.  On the other hand, the performance of $\BS$
becomes considerably bad if we underestimate $\gammaz$ and the number
of examples needed by this algorithm migh become huge.

\begin{center}
\begin{minipage}{0.47\textwidth}
\begin{center}
\psbox[width=\textwidth]{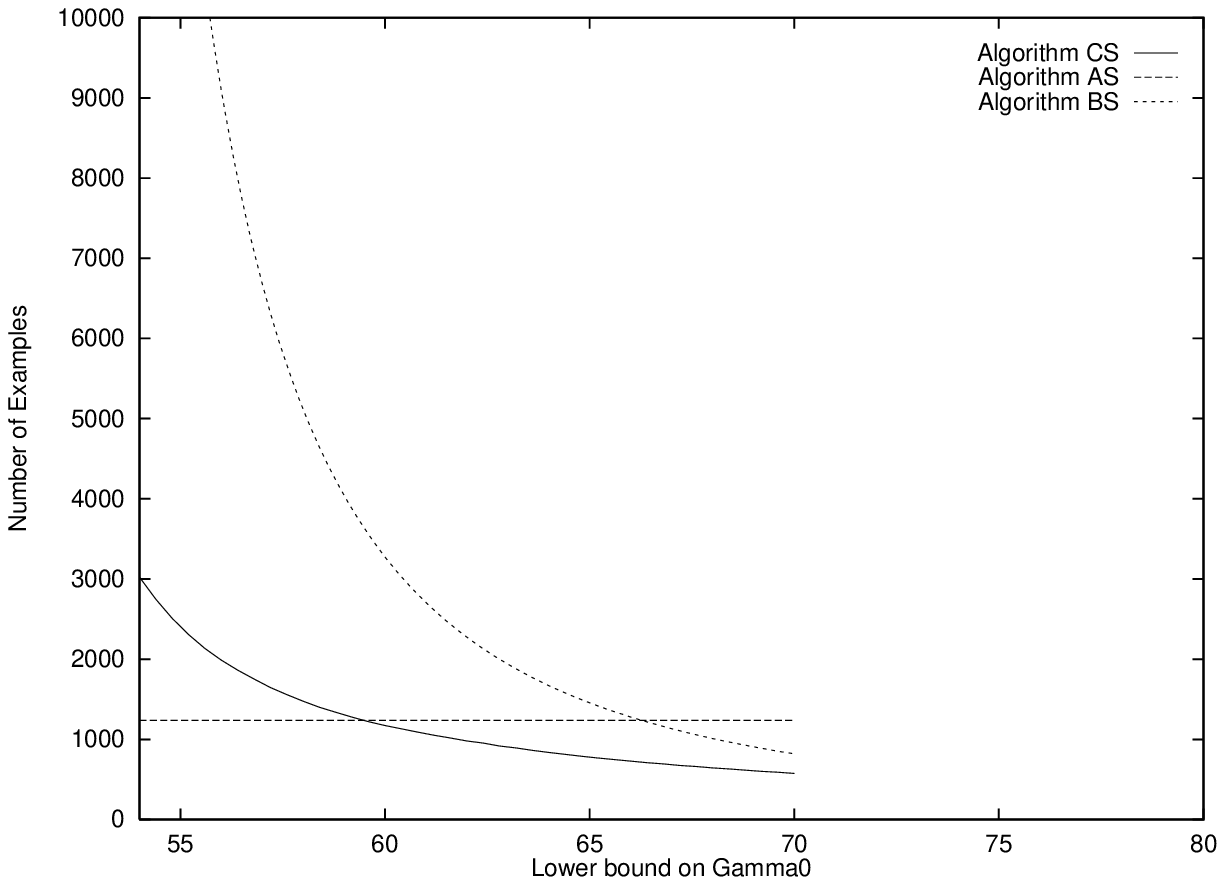}~\\
Figure 2:
$t$ vs.\ $\gamma$
($\gammaz=0.2$)
\end{center}
\end{minipage}~
\begin{minipage}{0.47\textwidth}
\begin{center}
\psbox[width=\textwidth]{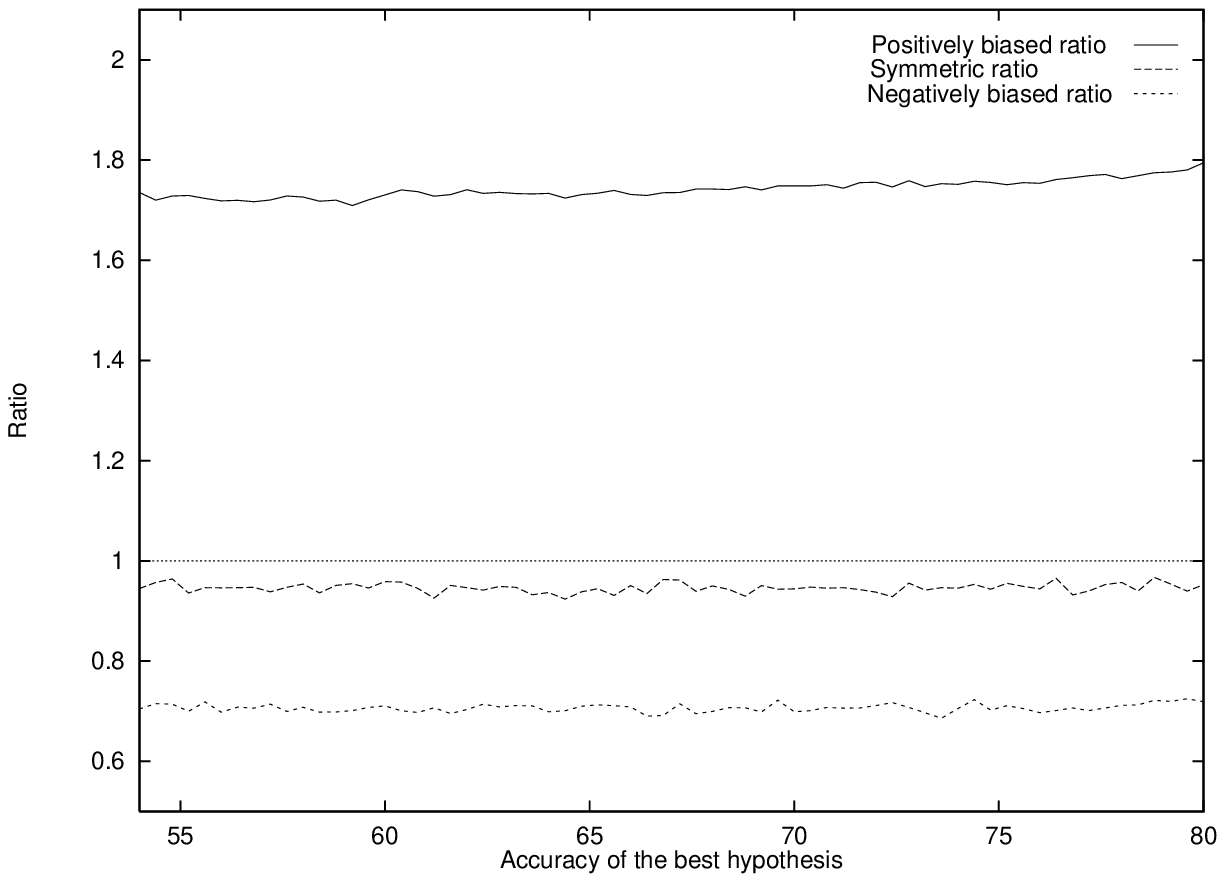}~\\
Figure 3:
$t/B\gammaz$ vs.\ $\gammaz$
\end{center}
\end{minipage}\\[2mm]
($t$ denotes the number of examples.)
\end{center}

\paragraphskip\noindent
\underline{3.~$\CS$: Constant $dec$ vs.\ Variable $dec$}

\noindent
For simplifying our theoretical analysis, we assumed that $dec$
(recall that $dec$ was $n'/n$) is constant $1/2$.  In fact, there are
two choices: either (i) to use constant $dec$, or (ii) to use variable
$dec$.  We investigate whether it affects the performance of the
algorithm $\CS$.  We verified that it does not affect at all the
reliability of $\CS$.  On the other hand, it affects the efficiency of
$\CS$, i.e., the number of examples needed by $\CS$.

Intuitively the following is clear: If the distribution of hypotheses
accuracy is symmetric (like in the above experiment), then the number
of successful hypotheses, at each step, is about $n/2$; thus,
$dec\approx1/2$, and the number of examples does not change between
(i) and (ii).  On the other hand, if most of the hypotheses are better
than $1/2$ (resp., most of the hypotheses are worse than $1/2$), then
the number of examples gets larger (resp., smaller) in (ii) than in (i).
We verified this intuition experimentally.
Figure~3
shows the ratio between
the number of examples and $B/\gammaz$
(which is always close to 1 if $dec=1/2$)
for three different distributions of hypotheses accuracy:
symmetric, positively biased, and negatively biased.
Thus,
when the distribution is negatively biased, which is the case in many
applications, we recommend to use the original $\CS$ with variable $dec$.

\paragraphskip\noindent
\underline{4.~$\AS$: $\eps$ vs. $\gammaz$, and the Theoretical Bound}

From the theoretical analysis of Theorem~\ref{theo:AStime}, we
obtained that the algorithm stops with high probability when $\eps$
becomes smaller than $\gammaz/4$.  On the other hand, to guarantee the
correctness of our algorithm (Theorem~\ref{theo:AS}), we just need to
conclude that $\eps$ is smaller than $\gammaz/2$.  This difference
gets reflected in our theoretical bound for the number of examples.
Our experiments (see Figure~4) showed that the number of examples is
much smaller than the theoretical bound.  The reason is that, in most
cases, the algorithm stops much before $\eps$ becomes as low as
$\gammaz/4$; it is more likely, that $\AS$ stops as soon as $\eps$
becomes slightly smaller than $\gammaz/2$.  Figure~5 reflect this
phenomenon; the final value of $\eps$ is closer to $\gammaz/2$ than
$\gammaz/4$.  (It is in fact on the $\gamma/2.38$ line.)  If we assume
that the final value of $\eps$ is about $\gammaz/2.38$ then, by using
the relation between $t$ and $\eps$, we can estimate the number of
examples as $4(2.38)^2\ln(3n/\delta)/(\chernoff\gammaz^2)$.

\begin{center}
\begin{minipage}{0.45\textwidth}
\begin{center}
\psbox[width=\textwidth]{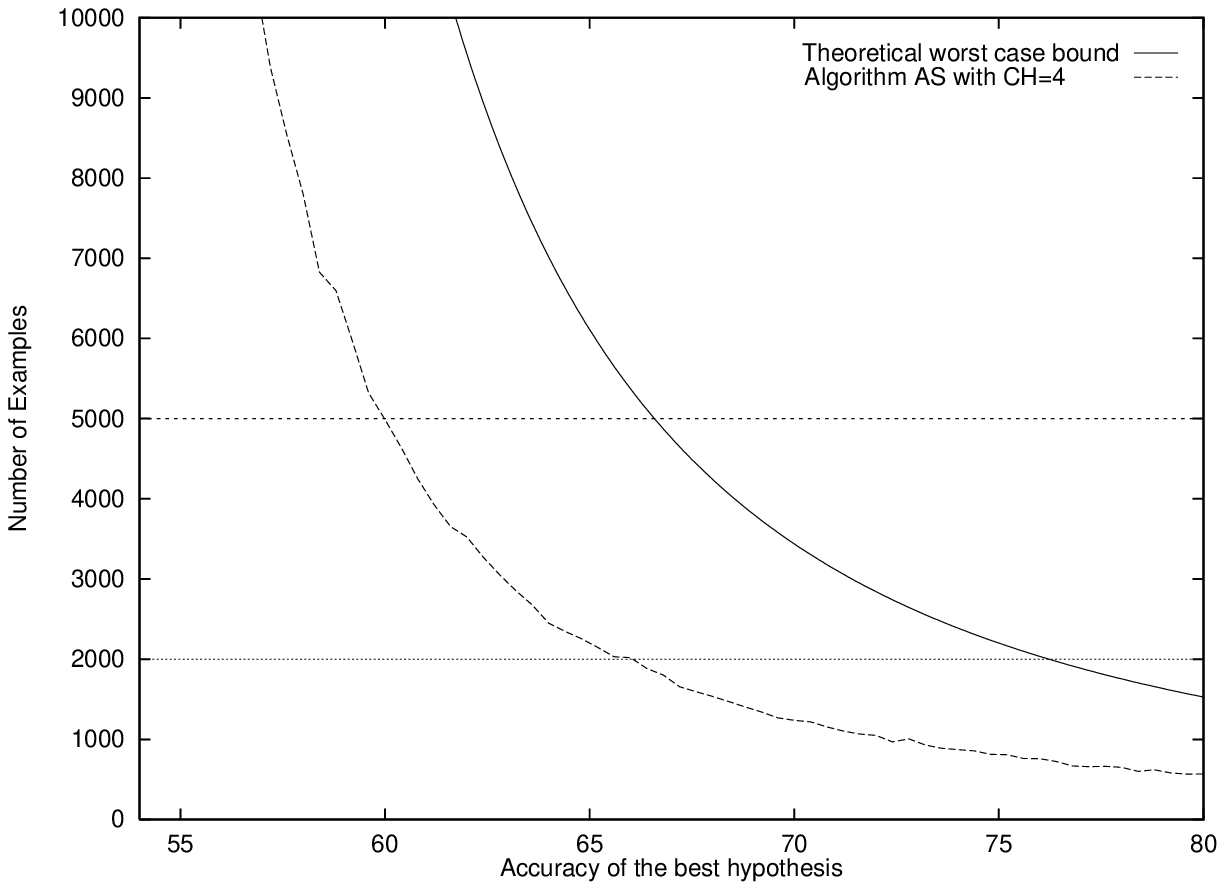}~\\
Figure 4:
$t$ and $\tAS$
\end{center}
\end{minipage}~
\begin{minipage}{0.45\textwidth}
\begin{center}
\psbox[width=\textwidth]{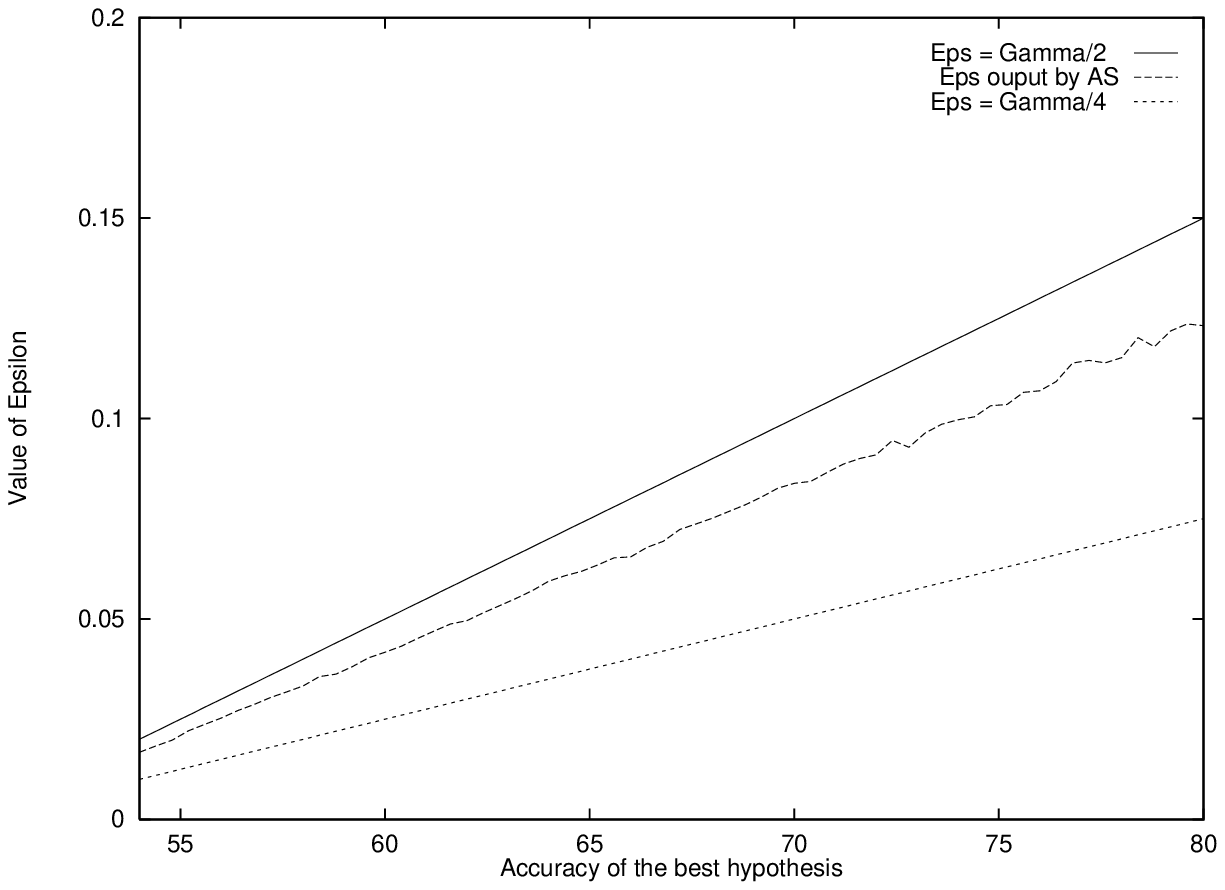}~\\
Figure 5:
the final value of $\eps$
\end{center}
\end{minipage}
\end{center}



\end{document}